\documentclass[journal]{IEEEtran}
\usepackage{amssymb}
\usepackage[algoruled,vlined,linesnumbered]{algorithm2e}
\usepackage{comment}
\usepackage{amsthm}

\theoremstyle{definition}

\theoremstyle{remark}

\usepackage{bm}
\usepackage{cite} 
\usepackage{amsmath}
\usepackage{float}
\usepackage{graphicx}
\usepackage{subfigure}
\usepackage{xcolor}
\usepackage{booktabs}

\usepackage{amssymb}
\usepackage{multirow}

\begin{document}
\title{Video Summarization through Reinforcement Learning with a 3D Spatio-Temporal U-Net}

\author{Tianrui~Liu, 
        Qingjie~Meng,
        Jun-Jie~Huang, 
        Athanasios~Vlontzos, 
        Daniel~Rueckert,~\IEEEmembership{Fellow,~IEEE,}
        and~Bernhard~Kainz~\IEEEmembership{Senior member,~IEEE,}
        }
        
\maketitle

\begin{abstract}
Intelligent video summarization algorithms allow to quickly convey the most relevant information in videos through the identification of the most essential and explanatory content while removing redundant video frames. In this paper, we introduce the 3DST-UNet-RL framework for video summarization. A 3D spatio-temporal U-Net is used to efficiently encode spatio-temporal information of the input videos for downstream reinforcement learning (RL). An RL agent learns from spatio-temporal latent scores and predicts actions for keeping or rejecting a video frame in a video summary. We investigate if real/inflated 3D spatio-temporal CNN features are better suited to learn representations from videos than commonly used 2D image features. Our framework can operate in both, a fully unsupervised mode and a supervised training mode.
We analyse the impact of prescribed summary lengths and show experimental evidence for the effectiveness of 3DST-UNet-RL on two commonly used general video summarization benchmarks. We also applied our method on a medical video summarization task. The proposed video summarization method has the potential to save storage costs of ultrasound screening videos as well as to increase efficiency when browsing patient video data during retrospective analysis or audit without loosing essential information.

\end{abstract}

\begin{IEEEkeywords}
Video summarization, Reinforcement learning, 3D convolutions, 3D U-Net, Medical video processing, Ultrasound.
\end{IEEEkeywords}

\IEEEpeerreviewmaketitle

\section{Introduction}
%
%
%
%
\IEEEPARstart{T}he amount of video data, including videos shared across social media, has been growing at an exponential rate. It is reported that for example approximately 300 hours of videos are uploaded to YouTube every single minute. This amount of information is impossible to process for humans. Currently, searching for desired content is mainly relied on textual key words and video thumbnails, thus search times often exceed genuine video utilization. This inhibits consumers and content creators alike.


There has been a steadily growing interest in machine learning techniques for video summarization. Early works adopt low-level or mid-level visual features to locate important segments of a video with a particular strategy such as clustering \cite{dynamic_cluster2013,SumMe_ECCV2014} and sparse dictionary learning \cite{sparsedictionary2011,SparseLatentDictionary_2014_CVPR}. Other methods rely on heuristic frame sampling mechanisms like static video storyboard summaries~\cite{li2001overview,Tianrui_vSumm2014} or frame-similarity-based dynamic video skimming~\cite{truong2007video}. However, most of these methods are not capable to model a temporal component that can provide additional contextual information for enhanced discriminative power~\cite{AnalyzingTemporal2018,spatiotemporal_action2019}.

Incorporating recurrent neural networks (RNNs) for sequence modeling has led to significant progresses in this field~\cite{LSTMvSumm2016,SUM-GAN_cvpr2017,hierarchical-lstm2017,HSA_RNN_cvrp2018}. Long short term memory (LSTM)-based methods using bi-directional LSTMs \cite{LSTMvSumm2016} as well as hierarchical LSTMs \cite{hierarchical-lstm2017,HSA_RNN_cvrp2018} have been proposed for sequence modeling in video summarization. 
Recently, Rochan~\textit{et al}.~\cite{FCSN_eccv2018} have demonstrated that it is possible to model the video summarization task as an element-wise segmentation problem using fully convolutional networks (FCNs). One of the most common FCN architectures are U-Nets~\cite{U-Net2015}, which have originally been proposed for image segmentation tasks where they exhibit good semantic image feature extraction abilities in the spatial domain~\cite{badrinarayanan2017segnet}. For video summarization, Rochan~\textit{et al}. proposed to utilize fully convolutional sequential network (FCSN)~\cite{FCSN_eccv2018} in the temporal domain for sequential modeling. As a segmentation tool, FCSN~\cite{FCSN_eccv2018} works for the video summarization problem by determining which parts of the video should be ``segmentated" as important. Compared to methods using recurrent models~\cite{LSTMvSumm2016,SUM-GAN_cvpr2017,HSA_RNN_cvrp2018}, FCSNs have the advantage of operating on the whole video, which provides the maximum amount of context and allows improved GPU utilization.
Encouraged by the success of FCSN~\cite{FCSN_eccv2018}, we propose a new 3D U-Net-based architecture to model both the spatial and the temporal dependencies among video frames. While the FCSN in~\cite{FCSN_eccv2018} takes 2D CNN features for each single frame and applies 1-dimensional (1D) convolutions along the temporal dimension for sequential modeling, we hypothesize that spatio-temporal features as well as full 3D convolutions are better suited for video data. 

In \cite{C3D_iccv2015,C3D_eccv2010,I3D_2017}, spatio-temporal CNN features have been used to improve the video action recognition task. However, to the best of our knowledge, a 3D spatio-temporal U-Net architecture using spatio-temporal features exploited for video summarization remains a rarely addressed problem.

In this paper, we propose a 3D spatio-temporal U-Net (3DST-UNet) using spatio-temporal features for video summarization. We accommodate the paradigm of Reinforcement Learning (RL) and exploit the combination of RL and 3DST-UNet. Our 3DST-UNet directly links with spatio-temporal feature extraction networks by taking as input 4-dimensional (4D) video features, encoding both spatial and temporal video information. 
A feature extraction CNN outputs spatio-temporal features of frame sequences and feeds them into the 3DST-UNet. The proposed 3DST-UNet exploits contextual information and maps the spatio-temporal video features effectively into a continuous latent space. Both the spatio-temporal features and the 3DST-UNet have the potential to encode spatio-temporal dependencies between video frames. The role of the RL agent is to learn policies which can maximize the accumulated reward of taking actions, where the actions are defined as whether to select or discard the current frame as a key frame.
The proposed 3DST-UNet model can be trained either in a supervised way or a fully unsupervised way, that is, the training process of our summarization network accommodates cases where little to no human annotations are available. 

To evaluate the effectiveness of the proposed method, we conduct comprehensive experiments using different training settings on public video summary datasets, \emph{i.e.}, SumME~\cite{SumMe_ECCV2014}, TVSum~\cite{TVSUM} as well as OVP and YouTube~\cite{OVP,VSUMM2011}. Furthermore, we extend our video summarization method for medical videos and take fetal screening with ultrasound imaging as an example. We demonstrate that our method achieves better performance than previous approaches proposed for video summarization on both general videos and medical videos.

Overall, the contribution of this paper is as follows:

(1) We propose 3DST-UNet-RL, a deep RL-based framework using a 3D spatio-temporal U-Net (3DST-UNet) for video summarization. To the best of our knowledge, this is the first approach to use a 3D U-Net with spatial-temporal features for video summarization.

(2) We introduce a 3D spatio-temporal U-Net (3DST-UNet) for sequential spatio-temporal video feature modeling. Compared RNN-based models, our 3DST-UNet directly links with spatio-temporal CNN features to exploit both spatial and temporal information.

(3) We provide comprehensive experiments and demonstrate that our method achieves better performance than previous approaches on two commonly used video summarization benchmarks. We also discussed the limitations of the current evaluation criteria for general video summarization and evaluate the video summary under various length constrains. 

(4) We extend our video summarization method for ultrasound scanning videos, where automated report generation with short but relevant video clips is desirable. We show experimental evidence for the effectiveness of our approach in medical video data from the clinical practice.


The rest of the paper is organized as follows: Section \ref{sec:literature} gives a review on the related works, Section \ref{sec:method} introduces the proposed 3DST-UNet-RL framework for video summarization, Section \ref{sec:results} shows the ablation studies of the proposed method and compares with other video summarization methods, and finally Section \ref{sec:conclusions} concludes the paper.

\begin{figure*}[th]
\begin{centering}
\center\includegraphics[width=1.98\columnwidth]{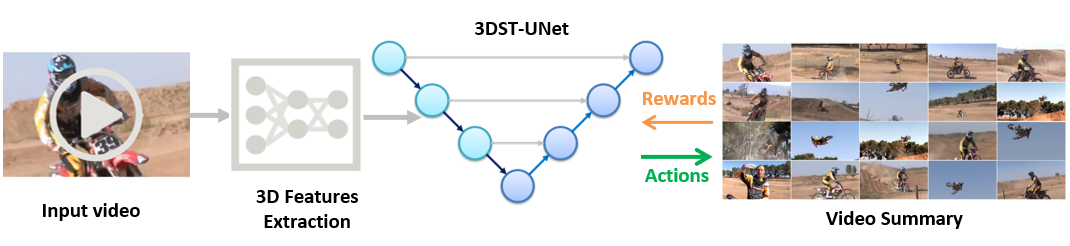}
\par\end{centering}
\caption{Overview of the proposed 3DST-UNet-RL method for video summarization. The framework consists of three main parts:a spatio-temporal CNN with 3D convolution for video features extraction, a 3D spatio-temporal U-Net (3DST-UNet) for sequence modeling, and an RL network which takes actions to select key frames for the summary set. Given an input video, the 3D feature extraction network first computes spatio-temporal feature and feeds them into a spatio-temporal U-Net for sequence modeling. Following, we make use of reinforcement learning to predict actions keeping/rejecting a frame for inclusion in a video summary.}
\label{fig:overview}
\end{figure*}
\section{Related Works}
\label{sec:literature}

\subsection{Video Summarization}

Video summarization methods have been explored in literature using either supervised learning or unsupervised learning. Supervised methods learn video summarization from manually labeled data consisting of videos and their corresponding user annotations as ground-truth, usually key frames that are subjectively perceived as important. 
In \cite{LSTMvSumm2016}, a LSTM-based key frame selection model is trained by minimizing the cross-entropy loss between the estimated key frames and the user annotated ground-truth key frames. In order to ensure diversity of the selected frames, an additional objective based on a determinantal point process (DPP) is used. Zhang \textit{et al}.~\cite{retrospective_vSumm2018} combine sequential models for the summary creation with a retrospective encoder, which maps the summaries to an abstract latent space. The work of \cite{unpaired_cvpr2019} learns a mapping function from a set of web videos to a set of summary videos via an adversarial process. There are also video summarization method using Generative Adversarial Network (GAN). 
In~\cite{SUM-GAN_cvpr2017}, a subset of representative key frames is selected by training a summarizer to minimize the distances between videos and a distribution of their summaries using generative adversarial networks as critics. Similarly, \cite{Cycle-SUM2019} aims to maximize the mutual information between a summary and video using an information-preserving metric, two trainable discriminators and a cycle consistent adversarial learning objective.

While deep neural networks (DNNs) have achieved significant improvements for video summarization performance, they impose a heavy burden on algorithm designers to collect a huge amount of labeled data for fully supervised DNN models. Therefore, unsupervised DNN methods are attractive and there have been pioneering works showing promising results. 
Some summarization methods provide weak supervision through additional cues such as images and videos from the web~ \cite{WeakSup_WebVideo_2017ICCV,webprior_2013,weaklyweb_eccv2018} and their accompanying category information \cite{KTS_2014,RLvSummBMVC2018} to improve performance. In ~\cite{weaklyweb_eccv2018}, a variational auto-encoder is applied for learning the latent semantics from web videos. In addition, an attention network for saliency estimation is used to improve the performance.

\subsection{Reinforcement Learning for Video Analysis}
The goal of reinforcement learning (RL) is to learn a good policy for the agent from experimental trials by maximizing expected future rewards. Reinforcement Learning (RL) has been succeeded in solving various vision tasks, such as visual tracking~\cite{RL_visual_track}, video face recognition \cite{RL_video_faceRecg}, video captioning~\cite{RL_video_caption}, and video object segmentation\cite{RL_video_segmentation}.

Recently, deep reinforcement learning (RL) has been explored for video summarization and videos fast-forwarding since these two tasks fit the narrative of RL well. Zhou and Qiao~\cite{RLvSumm_AAAI2018} propose a deep summarization network, which formulates the video summarization task as a sequential decision making process. The summaries are generated by predicting the probabilities of a given frame being a key-frame. The summary frames are then sampled based on this probability. In~\cite{RLvSummBMVC2018}, a Q-learning-based summarization network is explored to guide an artificial RL agent to use a recognizability reward. The reward is derived from a category classification network and the  summarization network relies on video-level category labels to address the summarization problem in a weakly supervised manner~\cite{RLvSummBMVC2018}. In ~\cite{RL_dpp_ECCV2018,FFNet_cvpr2018} RL has been used to fast-forward lengthy videos. A FFNet (i.e. FastForward Network) is proposed in  ~\cite{FFNet_cvpr2018} based on a Markov decision process to automatically decide the number of frames required for efficient fast-forwarding.

RL has also been applied in the field of medical image and video analysis. In ~\cite{vlontzos2019multiple,alansary2019evaluating}, RL strategies are successfully combined with spatio-temporal frame histories for video landmark detection. In ~\cite{liu_MICCAI_2020}, an RL based video analyse method is used to preserves essential information in ultrasound diagnostic videos. A diagnostic plane detection reward has been proposed which guides the agent to learn according to the clinical diagnostic standards. The features representing the video frames in~\cite{liu_MICCAI_2020} are extracted from 2D CNN networks and is further modelled using LSTM. Differently, we use the proposed 3DST-UNet to model the 3D CNN features for video summarization. 



\subsection{3D U-Net}
U-Net~\cite{U-Net2015} was originally proposed based on FCN for the biomedical image segmentation. Thereafter, the U-Net alike structures have been widely used in the field of image segmentation~\cite{U-Net2015}. U-Net resembles a fully convolutional network (FCN) structure and uses deconvolution to restore image size and feature. Different from FCN which the fusion operation during upsampling is direct feature addition, the U-Net upsampling process uses the concatenate operation to splicing the feature maps. Following concatenation is the feature map  deconvolvtion. The skip connection strategy directly utilizing shallow features.

Several studies \cite{3DUNet_MICCAI2016,3DUNet_brain_seg} have demonstrated that a 3D version of U-Net can produce better results than the 2D architectures. Çiçek et al.~\cite{3DUNet_MICCAI2016} proposed a 3D U-Net model that generate dense volumetric segmentations. It realizes 3D image segmentation by inputting a continuous 2D slice sequence of 3D images. The network structure is similar to U-Net, with one encoding path and one decoding path, each has four resolution levels. The encoder gradually reduces the spatial dimension by continuously merging the layers to extract feature information, and the decoder portion gradually restores the target detail and the spatial dimension. 

Our proposed architecture is based on UNet and is influenced by the fully convolutional sequential network in ~\cite{FCSN_eccv2018} which use 1-dimensional (1d) convolutional along the temporal dimension for sequential modeling. We extend the idea of (FCSN) ~\cite{FCSN_eccv2018} and design a 3D temporal U-Net (3DST-UNet) that takes spatio-temporal video features as input from a preceding spatio-temporal 3D CNN.

\section{Method}
\label{sec:method}


\subsection{Overview}
As illustrated in Fig.~\ref{fig:overview}, our video summarization framework consists of three main parts: a spatio-temporal CNN with 3D convolution for video feature extraction, a 3D Spatio-Temporal U-Net (3DST-UNet), and a RL agent's network. The video feature extraction network outputs spatio-temporal features from a sequence of video frames. These features are then combined and fed into the 3DST-UNet, which can further model both the spatial and temporal relationships within the video sequence. The 3DST-UNet is followed by a Sigmoid layer to produce scores for each frame of the input video. Based on these scores, the RL agent's network takes actions to decide whether or not to select a frame for the summary.

\subsection{Video Feature Extraction Network}

\begin{figure}[t]
\begin{centering}
\center\includegraphics[width=0.98\columnwidth]{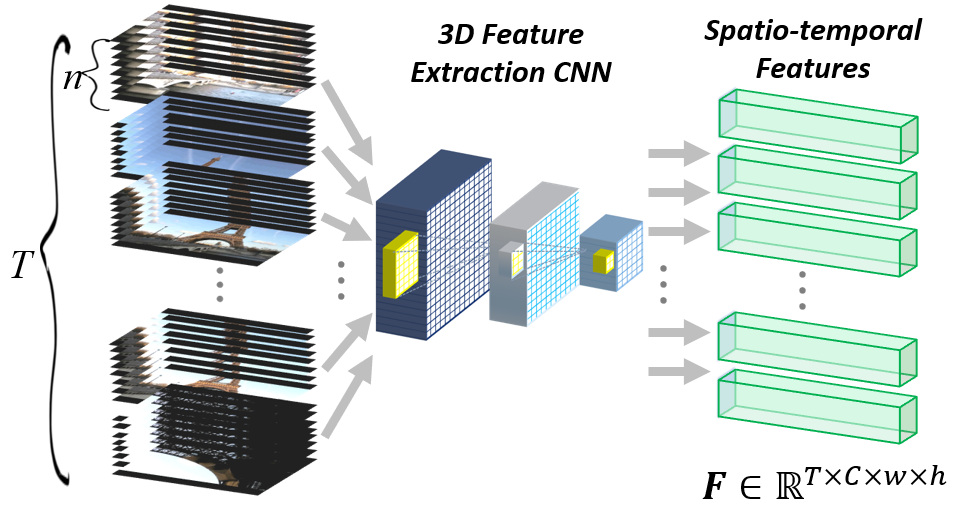}
\par\end{centering}
\caption{Sequential spatio-temporal video 3D feature extraction network. The input video of length $L = T \times n$ has been divided into $T$ segments of length $n$. The spatio-temporal video 3D feature extraction network takes $n$ frames at a time to compute video features of size $C \times w \times h$ and results in video feature $\bm{F} \in \mathbb{R}^{T\times C \times w \times h}$. }
\label{fig:featureNet}
\end{figure}


\begin{figure*}[t]
\begin{centering} 
\center\includegraphics[width=1.6\columnwidth]{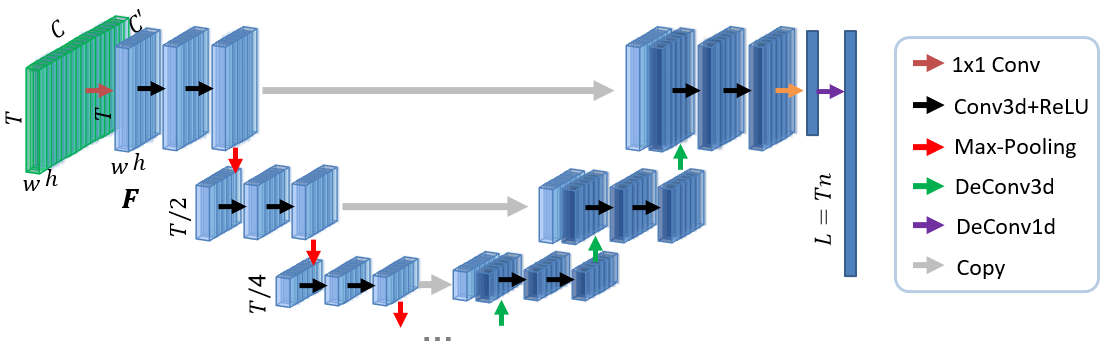}
\par\end{centering}
\caption{The proposed 3D Spatio-Temporal UNet (3DST-UNet). It takes 4-dimensional video feature $\bm{F} \in \mathbb{R}^{T \times C \times w \times h}$ as input where $C$, $w$, $h$ denotes the temporal, channel, spatial width, and spatial height dimensions of the feature, respectively. 3DST-UNet gradually extracts features at different temporal resolutions in the encoder part which are then successively recombined in the decoder part using upsampling operations to propagate context information to higher temporal resolution layers. 
}
\label{fig:3DST-UNet}
\end{figure*}

CNNs utilizing spatio-temporal convolutions are expected to suit video data better than 2D features because of their potential to directly encode spatio-temporal dependencies between frames. We investigate two options for video feature representation, \emph{i.e.,} spatio-temporal 3D features (ST3D) \cite{C3D_eccv2010} and inflated 3D (I3D) features \cite{I3D_2017}.
Specifically, we remove the last two fully connect layers of the backbone networks \cite{ResNet}. In this way, the extracted video feature representation is kept with the spatial dimensions $w$ and~$h$. 

\subsubsection{Spatio-temporal 3D Features}
As a direct solution, Spatio-temporal 3D (ST3D) features can generate spatio-temporal feature representations by allocating the third dimension of the input tensor as the temporal component. ST3Ds use 3D convolutional layers and have been proposed for video action recognition tasks~\cite{C3D_iccv2015,C3D_eccv2010,I3D_2017}. The preceding spatio-temporal CNN consists of three 3D convolutional layers which takes $n$ frames at a time to compute $T=L/n$ spatio-temporal feature vectors from the total $L$ frames for every $n$ neighboring frames. 


\subsubsection{Inflated 3D Features}
The performance of high-level computer vision tasks using real 3D convolutions can be limited when the pre-trained dataset is not large enough. In~\cite{I3D_2017}, Inflated 3D features (I3D) are proposed to make use of 2D CNNs with ImageNet~\cite{imagenet2009} pre-trained weights. I3D duplicates the 2D kernels along the axial direction to produce 3D kernels. We accommodate the inflated 3D convolution network to efficiently encoding the 3D video sequences. This allows our summarization framework to fully exploit 3D context information while re-purposing off-the-shelf deep 2D network structures and inheriting their large capacities to cope with image variances.
Similar to the ST3D features, the convolutional layers compute $T=L/n$ feature vectors for every $n$ neighboring frames. 

Hence, for both the ST3D and I3D features, the video feature representation is represented as $\bm{F} \in \mathbb{R}^{T\times C \times w \times h}$ where $C$, $w$, $h$ denotes the feature temporal, channel, spatial width, and spatial height dimensions, respectively. 
The block diagram of the 3D feature extraction network can be seen in Fig. \ref{fig:featureNet}.

\begin{figure*}[t]
\begin{centering}
\center\includegraphics[width=0.75\textwidth]{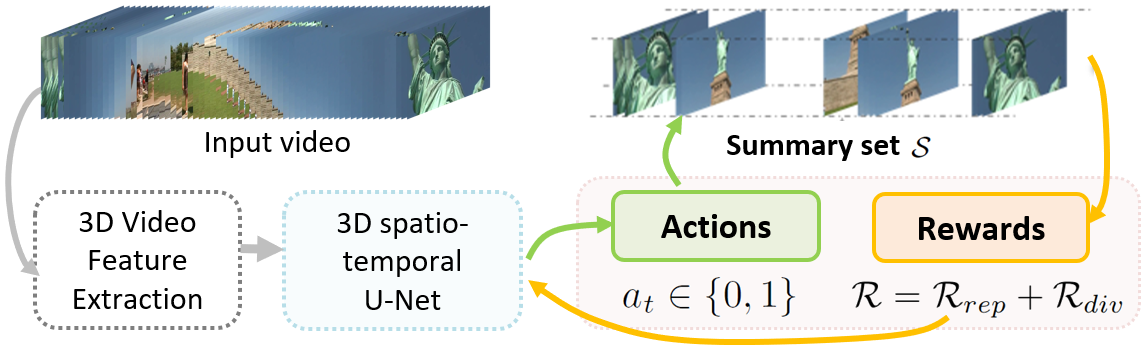}
\par\end{centering}
\caption{Reinforcement learning on the latent scores produced by the 3DST-UNet. With the latent scores from the 3DST-UNet, the reinforcement learning network selects actions $a_t$ on whether a frame should be selected into the summary set $\mathcal{S}$ or not by maximizing the rewards including the representativeness reward $\mathcal{R}_{\mathrm{rep}}$ and diversity reward $\mathcal{R}_{\mathrm{div}}$. 
}
\label{fig:RL}
\end{figure*}

\subsection{3DST-UNet for Sequential Spatio-temporal CNN Feature Modeling}

We propose to use a 3D Spatio-Temporal U-Net (3DST-UNet) to model the spatio-temporal video feature representation $\bm{F}$ in both, the temporal and the spatial domain for video summarization. 
The 3DST-UNet aims to generate latent scores for each video frame and takes features which encode both the spatial and the temporal video information as input. As such, the proposed 3DST-UNet is directly linked with the spatio-temporal feature extraction network for sequential modeling.

Different from the ordinary U-Net~\cite{U-Net2015} for semantic segmentation in the spatial domain ~\cite{badrinarayanan2017segnet}, the proposed 3D spatio-temporal U-Net (3DST-UNet) is proposed to be used for sequential modeling in both spatial and temporal domain. Also different from the fully convolutional sequential network (FCSN) which uses 1-dimensional convolutional along the temporal dimension, 3DST-UNet takes spatio-temporal video features from a preceding spatio-temporal 3D CNN as input. The 3DST-UNet for sequential modeling has the advantage of processing the entire video sequence at once. It does not employ recurrent or sequential processing, therefore can be driven in a single forward/backward pass during inference/training for sequences with variable length. Another merit of 3DST-UNet is that it is able to extract context information from different temporal resolutions and exploit the spatio-temporal information which are very helpful for video summarization.

The proposed 3DST-UNet is an end-to-end volumetric architecture consisting of an encoder, decoder stage, and up-sampling stage for video frame latent score generation. As illustrated in Fig.~\ref{fig:3DST-UNet}, the 3DST-UNet encoder generates features at different temporal resolutions and a decoder which successively fuses multi-resolution features and produces latent scores for each frame. 
Given the 4-dimensional (4D) video feature sequence of dimension $T \times C \times w \times h$, the encoder part of 3DST-UNet takes the 4D video feature $\bm{F} \in \mathbb{R}^{T \times C \times w \times h}$ as input; the decoder part outputs $L=T\times n$ latent scores where $L$ equals to the total number of frames in the input video, $T$ is the total number of spatio-temporal features fed into the 3DST-UNet. 


\subsubsection{Encoder} 
 We introduced squeeze-and-excitation blocks \cite{SENet_cvpr2018,liu_TIP_2020} to the 3D U-Net architecture. The 4D features maps (denoted as green box in Fig. \ref{fig:3DST-UNet}) are first compressed by a squeeze layer to compress the features through $1 \times 1$ convolutions. The channel-wise dimension $C$ is thereafter squeezed from $C$ to $C'$, i.e, 
 \begin{equation}
     \widehat{\bm{f}}_i = \bm{v}_i * \bm{F}, 
 \end{equation}
 where $\widehat{\bm{f}}_i$ is the $i$-th feature element of the output feature map $\bm{F} = \left[{\bm{f}}_{1}, \cdots, {\bm{f}}_{C_{in}}\right] \in \mathbb{R}^{T \times H \times W \times C_{in}}$, $\bm{v}_i$ is the $i$-th learned filter in the squeeze layers for $i=1,\cdots,C'$, and `$*$' denotes convolution.


 The encoder of 3DST-UNet consists of repeated application of 3D convolution, each followed by a rectified linear unit (ReLU) and a max pooling operation with stride 2 for downsampling. The dimension of feature channels are doubled at each downsampling stage. The 3D convolution operations of kernel size $3 \times 3 \times 3$ sliding along the channel, width and height dimensions to exploit both the spatial and the temporal context information. The pooling operations are performed along the temporal dimension to enable context information extraction from different temporal resolutions.

\subsubsection{Decoder} 

Every step in the decoding path consists of an upsampling of the feature map followed by a $2 \times 2 \times 2$ deconvolution that halves the number of feature channels. 3DST-UNet performs 3D deconvolution along the temporal dimension to upsample the features from lower temporal resolutions.
There are skip connections between the encoder and decoder features at the same temporal resolutions to pass information from the encoder part to the decoder part. The concatenated features are then processed using two $3 \times 3 \times 3$ convolutions (each followed by a ReLU). At the final stage, 3DST-UNet performs global average pooling and 1-dimensional deconvolution to transform the 4D feature into latent scores of length $L=T\times n$ for the downstream reinforcement learning.


\subsection{Reinforcement Learning of Latent Scores for Video Summarization}

 An RL value network takes as an input the representation of the environment state and the action choice of the agent. In our case, where the summarization task is interpreted as an RL decision making process, the RL takes input the frame latent scores from the 3DST-UNet and an action is taken at every frame on whether to include it in the summary set or not.

Our RL network receives spatio-temporal features and uses fully convolution layers with a Sigmoid activation function to output a probability score of each action. 
\textcolor{black}{By denoting the spatio-temporal feature sequence as $\{\bm{x}_t\}^L_{t=1}$,} the frame-level probability scores can be defined as $p_t=\text{Sigmoid}(\bm{W} * \bm{x}_l + {b})$, where $\bm{W}$ and ${b}$ are the parameters of the fully convolution layer.  

As illustrated in Fig. \ref{fig:RL}, the RL agent can take \textit{actions} on whether a frame should be selected in the summary set $\mathcal{S}$ according to the frame-level probability scores $p_t$. For the video summarization problem, the \textit{actions} are binary values, \emph{i.e.}, $a_t \in \{0,1\}$ where $a_t=1$ indicates the frame $t$ should be selected in the summary set $\mathcal{S}=\left\{s_{i}\left|a_{s_{i}}=1, i=1, \ldots,\right| \mathcal{S} |\right\}$ and vice versa. The frame probability scores $p_t$ are sampled from a Bernoulli distribution, \emph{i.e.}, $a_t \thicksim B(p_t)$.
The objective of the RL summarization network is to maximize the expected rewards: 
\begin{equation}
\mathcal{R} = \mathcal{R}_\text{rep} + \mathcal{R}_\text{div}, 
\label{two_rewards}
\end{equation}
where the representativeness reward $\mathcal{R}_\text{rep}$ and the diversity reward $\mathcal{R}_\text{div}$ evaluate the quality of the selected summary $\mathcal{S}$.

\subsubsection{The representativeness reward}
$\mathcal{R}_\text{rep}$ measures how well the generated summary can represent the original video. By maximizing $\mathcal{R}_\text{rep}$ for the original video, the temporal information across the entire video can be maximally preserved. The degree of representativeness of a video summary is formulated as a $k$-medoids problem~\cite{Gygli_vSumm2015}. The agent is encouraged to select a set of medoids such that the MSE between video frames and their nearest medoids is minimal, \emph{i.e.,} 

\begin{equation}
    \mathcal{R}_{\mathrm{rep}}=\exp \left(-\frac{1}{L} \sum_{t=1}^{L} \min _{t^{\prime} \in \mathcal{S}}\left\|\bm{x}_{t}-\bm{x}_{i}\right\|_{2}\right),
\end{equation}
where $\bm{x}_i$ is the spatio-temporal feature representation of a selected frame in the summary set $\mathcal{S}$.


\subsubsection{The diversity reward} $\mathcal{R}_\text{div}$ measures the dissimilarity between the selected frames for the summary video. It enforces the agent to select frames with different visual representations into the summary set $\mathcal{S}$. Thus, redundancy is kept small in $\mathcal{S}$.
$\mathcal{R}_\text{div}$ is computed as the pairwise frame dissimilarity, \emph{i.e.}, 

\begin{equation}
     \mathcal{R}_{\mathrm{div}}=\frac{1}{{\vert\mathcal{S}\vert  (\vert\mathcal{S}\vert-1)}} \sum_{t \in \mathcal{S}} \sum_{i \in \mathcal{S} \atop i \neq t} d\left(\bm{x}_{t}, \bm{x}_{i}\right),
\end{equation}
where $d(\cdot,\cdot)$ is the pair-wise dissimilarity of images in the feature space. In this paper, we adopt the cosine distance for measuring pair-wise dissimilarity.  


\subsection{Network Training}


 
\textcolor{black}{In our summarization network, the loss terms and reward terms are jointly optimized in an end-to-end manner.} In addition to the introduced award terms, we use two regularization loss terms $\mathcal{L}_\text{reg}^p$ and $\mathcal{L}_\text{reg}^b$ to regularize the properties of the selected summary set. 

The proportion regularization loss $\mathcal{L}_\text{reg}^p$ is used to penalize the selection of a large number of frames in the summary set $\mathcal{S}$:
\begin{equation}
    \mathcal{L}_\text{reg}^p = || \frac{1}{L} \sum_{t=1}^{L}  p_{t}-\epsilon ||^{2},
    \label{eq:proportionloss}
\end{equation}
where $\epsilon$ is a scalar controlling the proportion of the selected frames.

The binary regularization loss $\mathcal{L}_\text{reg}^b$ is to encourage the learned frame probability to be binary since the annotations in video summarization datasets are given with binary labels: 
\begin{equation}
    \mathcal{L}_\text{reg}^b
    =  \left( \frac{1}{T}\sum_{t=1}^{L} | p_{t}-0.5| \right)^{-1}.
    \label{eq:binaryloss}
\end{equation}

In this way, the total regularization loss is:

\begin{equation}
    \mathcal{L}_\text{reg}= \mathcal{L}_\text{reg}^p + \lambda \mathcal{L}_\text{reg}^b,
    \label{eq:regloss}
\end{equation}
where $\lambda$ controls the relative importance of the two loss terms. 

\subsubsection{Unsupervised learning} In the unsupervised case, the ground-truth user annotated scores are not available during training. The regularization loss and the reward terms can be used to regularize the properties of the selected summary and guide the learning of selecting key frames from the video sequence. The combined cost function for the video summarization network is formulated as 
\begin{equation}
\mathcal{L}_\text{uns} = \mathcal{L}_\text{reg}- \mathcal{R},
\label{eq:unsup_loss}
\end{equation}
where the subscript in $\mathcal{L}_\text{uns}$ indicates that the loss terms and reward terms are optimized in a fully unsupervised manner in the $\text{3DST-UNet}_{\text{unsup}}$ variant of our method.



\subsubsection{Supervised learning} In case user annotations of key frames are available, we can extend our 3DST-UNet-RL method to a supervised learning model with an objective that minimizes the distance between the predicted frames-wise importance scores and user annotated scores. We denote this supervised model as $\text{3DST-UNet}_{\text{sup}}$. 
Thus, the cost function for the supervised summarization model can be defined as 
\begin{equation}
\mathcal{L}_\text{sup} = \mathcal{L}_\text{pred} + \mathcal{L}_\text{reg}- \mathcal{R},
\label{eq:sup_loss}
\end{equation}
where $\mathcal{L}_\text{pred}$ is the mean squared errors (MSE) between the predicted frame-wise importance scores $p_{t}$ and ground-truth user annotated scores $p_{t}^*$, \emph{i.e.}, 
\begin{equation}
    \mathcal{L}_\text{pred} =\frac{1}{L} \sum_{t=1}^{L} \left\| p_{t}-p_{t}^*\right\|^{2}.
    \label{eq:mse_loss}
\end{equation}

The objective of the RL network is to train a video summarization agent for the optimal policy $\pi$ which indicates the actions to take to maximize the overall reward. 
The expected reward $\mathcal{J}(\theta)$ is $\mathcal{J}(\theta)= \mathbb{E}_{p_{\theta}\left(a|\pi\right)}[\mathcal{R}_\text{rep} +\mathcal{R}_\text{div}]$,
where $p_{\theta}\left(a|\pi\right)$ denotes the probability distribution over the actions of sequences. 

We follow~\cite{RLvSumm_AAAI2018} to compute the derivative of the objective function and approximate the gradient by taking the average of $k$ repeated episodes for each video while subtracting a constant value $c$. This is used to avoid high variance which would make it difficult for the network to converge. The constant $c$ is computed as the moving average of the previous rewards. The derivative of the expected reward $\mathcal{J}(\theta)$ can be expressed as:
\begin{equation}
    \nabla_{\theta} J(\theta)=\mathbb{E}_{p_{\theta}\left(a_{1: T}\right)}\left[ \sum_{t=1}^{T} (\mathcal{R}-c) \nabla_{\theta} \log \pi_{\theta}\left(a_{t} | h_{t}\right)\right].
    \label{derivative_reward}
\end{equation}


\subsection{Key Frames to Key Shots Conversion} 
\label{sec:summary_generalization}
Once we obtained frame-level importance scores via the deep RL network, we generate summary videos following existing protocols~\cite{LSTMvSumm2016,FCSN_eccv2018}. To transfer a key frame-based summary into a keyshot-based summary, the testing video is first segmented into shot intervals using kernel temporal segmentation (KTS)~\cite{KTS_2014}. The shot intervals containing at least one keyframe are marked as key shots. Finally, the knapsack algorithm \cite{TVSUM} is applied to select key shots with the highest averaging keyframe scores while keeping the duration of summaries below a threshold. 
The default length of the summary videos are restricted to be 15$\%$ duration of the original video for fair comparison with other video summarization methods. We also perform experiments on summaries generated with different summary length constraints to analyse the impact of prescribed summary lengths.

\section{Experiments}
\label{sec:results}

\subsection{Datasets}
\label{sec:dataset}
In order to test the performance of our proposed 3DST-UNet-RL method on both general video sets and medical videos, we perform experiments on widely used video summarization benchmarks, \emph{i.e.}, SumMe \cite{SumMe_ECCV2014}, TVSum \cite{TVSUM}, OVP and YouTube~\cite{OVP,VSUMM2011}. We also demonstrate the effectiveness of the video summarization method on a fetal screening ultrasound video dataset\cite{liu_MICCAI_2020}.

\subsubsection{The TVSum Dataset}
The Title-based Video Summarization dataset \cite{TVSUM} contains 50 videos of various genres (e.g., news, documentary, egocentric) and 1,000 annotations of shot-level importance scores (20 user annotations per video). The duration varies from 2-10 minutes.

\subsubsection{The SumMe Dataset}
The {SumMe} dataset \cite{SumMe_ECCV2014} consists of 25 videos, each annotated with at least 15 human-annotated summaries. The duration of videos varying from 1.5–6.5 minutes.
The data consists of videos, annotations and source code for standardized evaluation, which we also use in our experiments.

\subsubsection{OVP and YouTube}
We utilize additional videos as augmentation datasets to alleviate overfitting as in \cite{LSTMvSumm2016,FCSN_eccv2018}.
The OVP \cite{OVP,VSUMM2011} and YouTube datasets \cite{VSUMM2011} are constructed for keyframe-based video summarization. OVP contains 50 videos downloaded from the Open Video Project. The duration of each video is 1–4 min. YouTube contains 50 videos collected from the YouTube website of duration 1–10 minutes. Both of OVP and YouTube are provided with five key-frame based summaries annotated by human. 

\subsubsection{The Ultrasound Dataset}
The ultrasound dataset are screen capture video recordings from fetal screening ultrasound examinations. There are 50 videos of 13-65 minutes length in our dataset from 50 different patients acquired between 24-30 weeks of gestation. The videos have been acquired and labelled during routine screenings according to the guidelines in the UK National Health Service (NHS) FASP handbook~\cite{FASP2015}.
The feature extraction network is trained on annotations indicating the type of standard ultrasound diagnostic plane. From all available FASP planes we have selected Brain (Cerebellum), Brain (Ventricle), Profile, Lips, Abdominal, Kidneys, Femur, Spine (Coronal), Spine (Sagittal), 4‐Chamber (4CH) cardiac view,  3 Vessel View (3VV)  cardiac view, Right Ventricular Outflow Tract (RVOT), Left Ventricular Outflow Tract (LVOT) as the most frequent exemplars. For ultrasound video summarization, we take the freeze-frame images which are saved by the sonographers during the scan as the ground-truth key frames.

\subsection{Experimental Settings} We implement the 3DST-UNet-RL method in PyTorch based on the architecture of the DSN~\cite{RLvSumm_AAAI2018} network. We take 3D ResNet50 architecture~\cite{3DResNet_2018} as the backbone network, which is a good compromise between performance and computational complexity. ResNet50 has shown competitive performance against other deeper architectures, for example for the task of action recognition. 

The experiments were run on a single TITAN RTX GPU. 
Stochastic Gradient Descent with momentum 0.9 and weight decay of $10^{-6}$ is used to train the 3DST-UNet-RL model. The initial learning rate was set to $10^{-5}$ for SumMe and $10^{-6}$ for TVSum, and was subsequently reduced by a factor of $0.5$ for every 30 epochs. We set $\lambda = 0.01$ and $\epsilon =0.5$ for Eq.~(\ref{eq:regloss}). For ST3D and I3D, the squeeze layer of in 3DST-UNet compress the features $\bm{F}$ through $1 \times 1$ convolutions from $C=2048$ to $C'=32$.
 
The supervised learning model relies on the use of a single ground-truth summary to compute the prediction loss. For ultrasound dataset, the ground-truth key frame are freeze-frame images which are saved by a single sonographer. For SumMe and TVSum dataset, however, there are multiple human-annotated summaries. Therefore, we follow prior works \cite{vSumm_NIPS2014,LSTMvSumm2016,FCSN_eccv2018} to generate an ``oracle'' summary \cite{vSumm_NIPS2014} for SumMe and TVSum dataset that maximally agrees with all annotators for each video. During training, the ``oracle'' summary is served as the ground-truth frame importance scores.

\subsection{Experimental Results}

\begin{table}[t]
\center
\caption{Comparisons of F1 scores using unsupervised approaches on the SumMe and TVSum datasets. (The best scores are in bold.)}
\begin{tabular}{lcc}
\toprule
Method                  & TVSum  &  SumMe    \\ \hline \hline
TVSum \cite{TVSUM}    & 50.0  & 36.0   \\ \hline
$\text{SUM-GAN}_\text{unsup}$ \cite{SUM-GAN_cvpr2017}   & 51.7 & 39.1    \\ \hline
Backprop-Grad \cite{WeakSup_WebVideo_2017ICCV}     & 52.7  & -   \\ \hline
$\text{SUM-FCN}_\text{unsup}$   \cite{FCSN_eccv2018}     & 52.7 & 41.5   \\\hline
$\text{UnpairedVSN}_\text{unsup}$~\cite{unpaired_cvpr2019}  &53.6 & 44.8   \\\hline
SASUM~\cite{SASUM_vSumm_AAAI2018}     & 53.9 & 40.6 \\ \hline
$\text{DR-DSN}_\text{unsup}$   \cite{RLvSumm_AAAI2018}  & {57.6} & 41.4     \\ \hline
$\text{DR-DSN}_\text{sup}$ (baseline)  & {56.7} &  43.2 \\ \hline
$\text{3DST-UNet}_\text{unsup}$(ours) & \textbf{58.1} &  \textbf{44.6}\\ 
\bottomrule
\end{tabular}

\label{tab:SoA-unsup}
\end{table}

We performed experiments on five different splits of training and testing subsets on SumMe and TVSum datsets. The percentages for training and testing subsets are $80\%$ and $20\%$ for both TVSum and SumMe datasets. The averaged F1 scores of the five splits are computed for both unsupervised and supervised learning paradigms.

Following the protocols in \cite{Gygli_vSumm2015,SumMe_ECCV2014,TVSUM}, we compute the precision ($P$) and recall ($R$) as well as their harmonic mean $F1$-score against the user summary for evaluation, \emph{i.e.}, 
\begin{equation} 
F=\frac{(2P\times R)}{(P+R)},
\end{equation}
where $P$ and $R$ are computed according to the temporal overlap, \emph{i.e.}, $|\mathcal{A} \cup \mathcal{B}|$ between a user annotated summary $\mathcal{A}$ and a network predicted summary $\mathcal{B}$:
\begin{equation} 
P =\frac{\mathcal{A} \cup \mathcal{B}}{|A|}, \qquad R =\frac{\mathcal{A} \cup \mathcal{B}}{|B|},
\end{equation}
where $|\mathcal{A}|$ and $|\mathcal{B}|$ denotes the duration of summary $\mathcal{A}$ and $\mathcal{B}$, respectively.


\subsubsection{Comparison to State-of-the-Arts}

We compare our proposed $\text{3DST-UNet}_{\text{unsup}}$ with state-of-the-art unsupervised video summarization methods including TVSum \cite{TVSUM}, SUM-GAN~\cite{SUM-GAN_cvpr2017}, Backprop-Grad~\cite{WeakSup_WebVideo_2017ICCV}, SASUM~\cite{SASUM_vSumm_AAAI2018}, SUM-FCN~\cite{FCSN_eccv2018} and DR-DSN \cite{RLvSumm_AAAI2018}. The results are given in Table~\ref{tab:SoA-unsup}. 

The supervised training result of the proposed method $\text{3DST-UNet}_{\text{sup}}$ are compared in Table \ref{tab:SoA-sup}. 
$\text{SUM-GAN}_\text{sup}$~\cite{SUM-GAN_cvpr2017}, $\text{DR-DSN}_\text{sup}$~\cite{RLvSumm_AAAI2018}, and $\text{SASUM}_\text{sup}$~\cite{SASUM_vSumm_AAAI2018} are extended from the unsupervised version by adding the discriminative loss of generated summaries and human-annotated summaries.

From the experimental results in Table~\ref{tab:SoA-unsup} and \ref{tab:SoA-sup}, our method outperforms the recurrent sequence modeling approach, \emph{i.e.}, VSUMM~\cite{VSUMM2011}, dppLSTM~\cite{LSTMvSumm2016} and SUM-GAN~\cite{SUM-GAN_cvpr2017}. 
To be noted, the $\text{DR-DSN}_\text{sup}$ (baseline) in Table~\ref{tab:SoA-unsup} and \ref{tab:SoA-sup} indicated the results that we obtained by using the same training/testing data splits as in our methods. Under the strictly fair comparison setting, our performance surpass the comparison method not only for supervised model but also for unsupervised model on both two testing datasets.

\begin{table}[t]
\center
\caption{Comparisons of F1 scores using supervised approaches on the SumMe and TVSum datasets. (The best scores are in bold.)
}
\begin{tabular}{lcc}

\toprule
Method                  & TVSum &  SumMe  \\ \hline \hline
LSTM~\cite{LSTMvSumm2016}  & 54.2  & 37.6    \\ \hline
dppLSTM~\cite{LSTMvSumm2016}  & 54.7  & 38.6    \\ \hline
$\text{SUM-GAN}_\text{sup}$~\cite{SUM-GAN_cvpr2017}   & 56.3 & 41.7     \\ \hline
$\text{SASUM}_\text{sup}$~\cite{SASUM_vSumm_AAAI2018}     & \textbf{58.2} & 45.3  \\ \hline
$\text{SUM-FCN}_\text{sup}$~\cite{FCSN_eccv2018}     & 56.8 & \textbf{47.5}  \\ \hline
$\text{UnpairedVSN}_\text{sup}$~\cite{unpaired_cvpr2019}     &55.6 & \textbf{47.5}  \\\hline
$\text{DR-DSN}_\text{sup}$~\cite{RLvSumm_AAAI2018}  & \textbf{58.1} & 42.1  \\ \hline
$\text{DR-DSN}_\text{sup}$ (baseline)  & {57.2} & 45.7   \\ \hline
$\text{3DST-UNet}_\text{sup}$(ours) & \textbf{58.3} & \textbf{47.4}  \\ 
\bottomrule
\end{tabular}
\label{tab:SoA-sup}
\end{table}


\subsection{Ablation Study}
\label{Ablation}
We first conduct ablation studies on the public video dataset \cite{TVSUM} to investigate (i) the effectiveness of spatio-temporal 3D CNN features, (ii) the performance of 3DST-UNet for sequential modeling compared to bi-directional LSTM (Bi-LSTM), and (iii) the effectiveness of the reward terms.

For all the experiments in this section, we use a single split of training and testing videos on the TVSum dataset. We compared the proposed 3DST-UNet with the Bi-LSTM network applied in our baseline method \cite{RLvSumm_AAAI2018} in terms of the sequential model performance. The performance has been evaluated using 2D, Inflated 3D (I3D) and spatio-temporal 3D (ST3D) features and has been given in Table \ref{tab:ablation}.

\subsubsection{Bi-LSTM vs. 3DST-UNet}
The 3DST-UNet and Bi-LSTM in Table~\ref{tab:ablation} indicates the summarization networks using our 3DST-UNet for sequential modeling and the baseline method \cite{RLvSumm_AAAI2018} uses a bi-directional LSTM network topped with a fully connected layer, respectively. For both 3DST-UNet and the Bi-LSTM model, a Sigmoid function is used to predict frame probability scores. 

As shown in Table~\ref{tab:ablation}, the 3DST-UNet outperforms the Bi-LSTM model on sequential estimation of importance scores using both 2D and 3D video features. These comparison results conform to the conclusion that the proposed 3DST-UNet can encode spatio-temporal information of the input videos more efficiently for the video summarization task.

\subsubsection{2D CNN features vs Spatio-temporal 3D features}

The 2D CNN features are extracted from the \textit{pool}5 layer of a GoogleNet \cite{GoogleNet} pre-trained with ImageNet \cite{imagenet2009} and are of dimension 
1024 which is identical to that in \cite{LSTMvSumm2016,FCSN_eccv2018,RLvSumm_AAAI2018}. The backbone networks of 2D features are pre-trained with ImageNet, while the backbone networks of I3D are pre-trained on the Kinetics 400 database~\cite{kinetics_dataset_2017}.
The ST3D features and I3D features takes $n=16$ frames at a time to compute representation features for the input video.
In order to fit the I3D and ST3D features into the Bi-LSTM network for comparison, the spatial dimension of the features are pooled into 1 by 1, that is $w=h=1$ for $\mathbb{R}^{T\times C \times w \times h}$. 

As compared in Table~\ref{tab:ablation}, the performance of using the I3D and the ST3D features are better than that of using 2D features, both with the Bi-LSTMs model and the 3DST-UNet model. These results show evidence that spatio-temporal 3D features have stronger representation ability for video sequential modeling.

\begin{table}[t]
\center
\caption{Ablation study comparing the proposed 3DST-UNet with the bi-directional LSTM network applied in our baseline method \cite{RLvSumm_AAAI2018} in terms of the sequential model performance. The performance has been evaluated using 2D, Inflated 3D (I3D) and spatio-temporal 3D (ST3D) features. (The best scores are in bold.)
}
\begin{tabular}{l ccc ccc}
\toprule
Method  & \multicolumn{3}{c}{Supervised} & \multicolumn{3}{c}{Unsupervised} \\ \hline \hline
Feature & 2D        & I3D     & ST3D      & 2D        & I3D       & ST3D      \\ \hline
Bi-LSTM     & 50.6     & 51.1    & 52.9    & 51.8     & 51.6     & 53.0   \\ 
Our  & 52.1     & 52.8    & \textbf{54.1}    & 52.3     & 52.7     & \textbf{53.6}   \\ 
\bottomrule
\end{tabular}
\label{tab:ablation}
\end{table}

\begin{table}[tp]
\center
\caption{F1 scores of the video summarization results on the TVSum dataset using an unsupervised and a supervised learning paradigm with ($\checkmark$) or without ($\times$)  $\mathcal{R}_{\text {rep}}$ and $\mathcal{R}_{\text {div}}$ reward terms. (The best scores are in bold.)}
\begin{tabular}{cccc}
\toprule
\multicolumn{2}{c}{Rewards} &
\multicolumn{2}{c}{Learning paradigm} \\
\hline
$\mathcal{R}_{\text {rep}}$ &  
$\mathcal{R}_{\text {div}}$  & 
Unsupervised & 
Supervised   \\ 
\hline \hline
$\times$   &  
$\times$   &   
-   &      
56.3    \\ 
\checkmark &  
$\times$  &   
55.1    &   
\textbf{56.9}       \\ 
$\times$      & 
\checkmark &   
54.5   &       
56.1      \\ 
\checkmark & 
\checkmark &     
\textbf{55.8}     &      
56.6         \\ 
\bottomrule
\end{tabular}

\label{tab:w_o_award}
\end{table}

\subsubsection{The effectiveness of reward(s)}

We further conduct ablation experiments by dropping the rewards terms to analyze the effect of the rewards using for RL learning. According to Eq. (\ref{eq:sup_loss}) and Eq. (\ref{two_rewards}), $\mathcal{L}_\text{sup} = \mathcal{L}_\text{pred} + \mathcal{L}_\text{reg}-(\mathcal{R}_\text{rep} +\mathcal{R}_\text{div})$.
By dropping both of the two reward terms, the model is trained using the regression loss $\mathcal{L}_\text{reg}$ as well as the MSE loss $\mathcal{L}_\text{pred}$ between the ground-truths and the predictions in a supervised manner. 
The results of ablation experiments on the effectiveness of reward terms are compared in Table~\ref{tab:w_o_award}. For the supervised learning settings, the F1 score increases by 0.6 by adding $\mathcal{R}_{\text {rep}}$. This indicates that the usage of RL can further enable the model to predict higher quality summaries than using only the supervised loss. 
It is worth noting that the supervised model implicitly achieves diversity to some extend by directly learning from the human ground-truth annotations. Hence, it is reasonable that the performance of using $\mathcal{R}_{\text {div}}$ is not better than the pure supervised model without using rewards.

\begin{figure}[t]
\begin{centering}
\center\includegraphics[width=0.88\columnwidth]{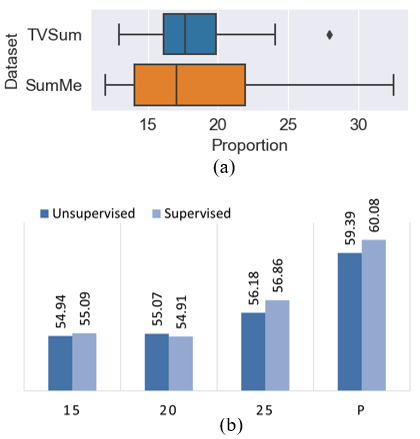}
\par
\end{centering}
\caption{(a) Box plot of the proportion of important frames given by human annotators in TVSum and SumMe datasets. (b) Comparison of F1 scores with four different summarization length constrains: $l=15\%$, $l=20\%$, and $l=25\%$ and $l=P$, where $P$ equals to proportion of important frames in the ground-truth for each video. }
 \label{fig:res_propotion}
\end{figure}


\subsection{Augmented and Transfer Data Settings}

To further analyze the results for our model from the main paper, we follow the prior works \cite{LSTMvSumm2016,FCSN_eccv2018,unpaired_cvpr2019} to utilize augmentation videos from OVP~\cite{OVP,VSUMM2011} and YouTube~\cite{VSUMM2011} as supplementary of the main SumMe and TVSum datasets.


\subsubsection{Canonical} 
The {canonical} setting is the standard form where the training
and testing sets are from the same dataset, as in our paper.

\subsubsection{Augmented} 
In the {augmented} setting, for a given dataset, we randomly leave $20\%$ of for testing, and augment the remaining $80\%$ with the other three datasets to
form an augmented training dataset. 

\subsubsection{Transfer} 
In the {transfer} setting, for a given dataset, we use the other three datasets for training and testing the learned models on the dataset. 


Table~\ref{tab:aug_trans_Summe} and Table~\ref{tab:aug_trans_TVSUM} show the performances of different methods in the Canonical (C), Augmented (A) and Transfer (T) settings on both TVSum and SumMe datasets. 

As we can see from the results, under the \textit{Augmented} setting, our proposed method has been further improved for both the supervised model and the unsupervised model. It is also noticed that, the supervised model does not improve as much as the unsupervised ones. This may due to the usage of “oracle” as the ground-truth frame importance score for supervised training. Although the “oracle” summary maximally agrees with all annotators for each video, however, as has been argued in \cite{vsumm_unsup_atten_GAN} and \cite{rethinking2019}, due to the highly-subjective of general video summarization task, it is not guaranteed that the inconsistent annotations from multiple users can fully explore the learning potential of a supervised model.


For the \textit{Transfer} setting, the the surprisingly good performance, as has also been reported in \cite{LSTMvSumm2016}, suggests a high correlation of the domains for the training datasets and the test dataset.
These results are encouraged for the cases that little to no human annotations are available.

\subsection{Results with Different Summary Lengths}
The above experimental results are conducted with a constant summary length constraint of $15\%$, \emph{i.e.}, the length of the summary videos are restricted to be shorter than $15\%$ of the input video length. Nevertheless, as we investigated in our experiments, the proportion of the important frames labeled by human annotators are not restricted with this threshold. We provide the distribution of summary length in the SumMe and TVSum datasets in Fig.~\ref{fig:res_propotion}(a). As evident, the proportion of the important frames labeled by human annotators can be as large as $30\%$. 

We believe that the video summary performance will be impacted if we restrict the summary length to be $ 15\%$ for input videos which have a ground-truth summary length of being longer than $15\%$. The video summarization network is forced to select only short video intervals of relatively high importance, instead of the intervals with highest scores. This will degrade the summary performance especially for the supervised models. 
\begin{table}[t]
\center
\caption{Performances of different methods in the Canonical (C), Augmented (A) and Transfer (T) settings on the TVSum datasets.}
\begin{tabular}{lccc}
\toprule
Method                  & C & A & T    \\ \hline \hline
LSTM$_\text{sup}$~\cite{LSTMvSumm2016}       & 54.2          & 57.9        & 56.9 \\ \hline
dppLSTM$_\text{sup}$~\cite{LSTMvSumm2016}    & 54.7          &59.6           &58.7    \\ \hline
$\text{SUM-FCN}_\text{unsup}$~\cite{FCSN_eccv2018}& 52.7 & - & 52.5 \\ \hline
$\text{SUM-FCN}_\text{sup}$~\cite{FCSN_eccv2018}  & 56.8 & 59.2 &58.2  \\ \hline
$\text{UnpairedVSN}_\text{unsup}$~\cite{unpaired_cvpr2019}&55.6 & - & 55.7\\\hline
$\text{3DST-UNet}_\text{unsup}$(ours) & 58.3 & 58.4 & 58.0 \\ \hline
$\text{3DST-UNet}_\text{sup}$(ours) & 58.3 & 58.9 & 56.1 \\ 

\bottomrule
\end{tabular}
\label{tab:aug_trans_Summe}
\end{table}

\begin{table}[t]
\center
\caption{Performances of different methods in the Canonical (C), Augmented (A) and Transfer (T) setting on the SumMe datasets.}
\begin{tabular}{lccc}
\toprule
Method                  & C & A & T    \\ \hline \hline
LSTM$_\text{sup}$~\cite{LSTMvSumm2016}       & 37.6   & 41.6  & 40.7 \\ \hline
dppLSTM$_\text{sup}$~\cite{LSTMvSumm2016}    & 38.6  &42.9  &41.8 \\ \hline

$\text{SUM-FCN}_\text{unsup}$~\cite{FCSN_eccv2018}& 41.5 & - & 39.5 \\ \hline
$\text{SUM-FCN}_\text{sup}$~\cite{FCSN_eccv2018}  & 47.5 & 51.5 &44.1  \\ \hline
$\text{UnpairedVSN}_\text{sup}$~\cite{unpaired_cvpr2019}&47.5 &  -& 41.6 \\\hline
$\text{3DST-UNet}_\text{unsup}$(ours) &44.6  & 49.5 & 45.7 \\ \hline
$\text{3DST-UNet}_\text{sup}$(ours) & 47.4 & 49.9 & 47.9 \\ 

\bottomrule
\end{tabular}
\label{tab:aug_trans_TVSUM}
\end{table}

\begin{figure*}[ht]
\begin{centering}
\center\includegraphics[width=2.0\columnwidth]{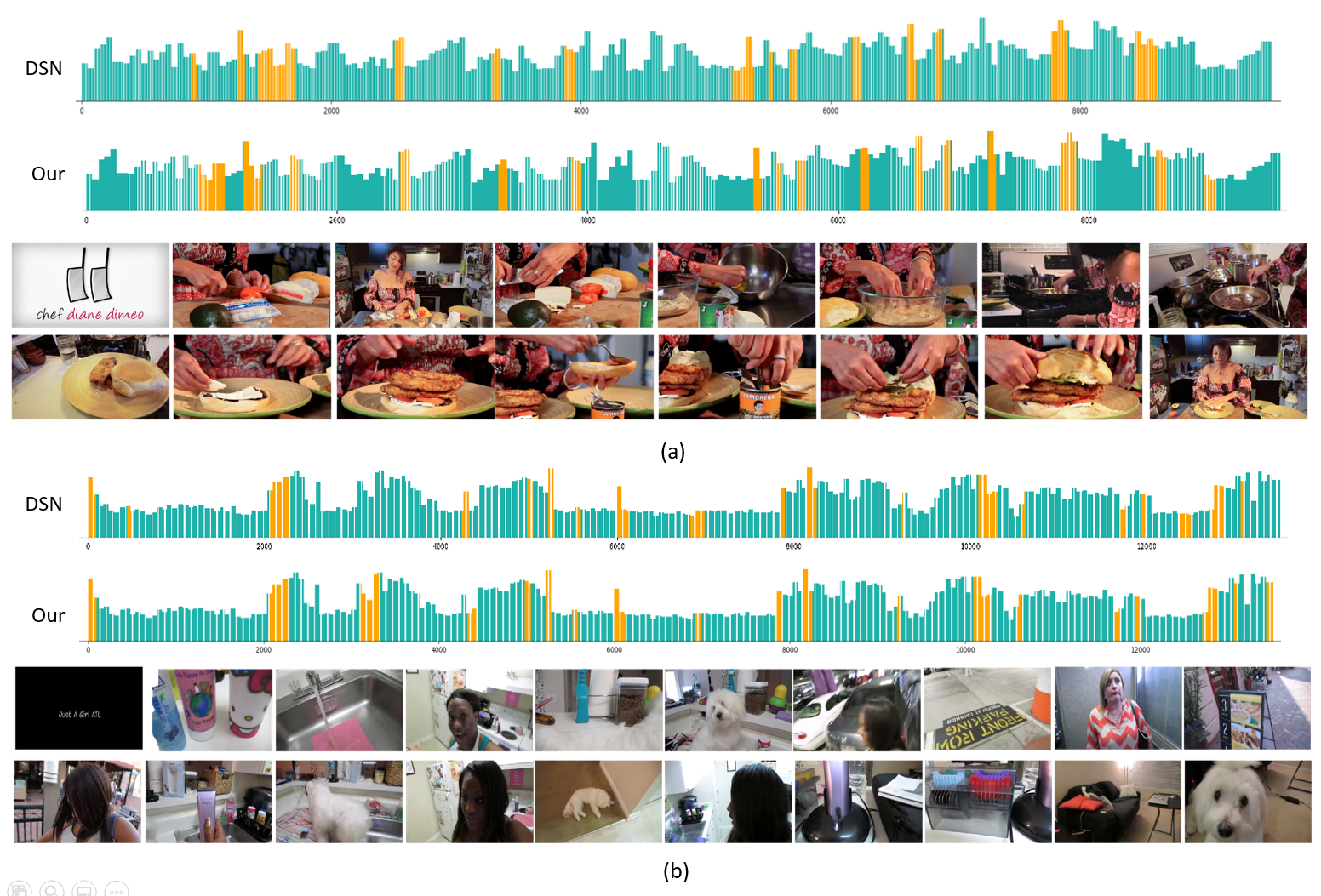}
\par
\end{centering}
\caption{Qualitative video summarization result of an exemplar video from the TVSum dataset. In the bar charts, the light-green bars correspond to ground-truth importance scores; the orange bars mark the intervals that have been selected by the proposed 3DST-UNet-RL and the baseline method DSN~\cite{RLvSumm_AAAI2018}, respectively. 
}
\label{fig:quality_eg}
\end{figure*}

To mitigate this problem, we perform experiments on summaries generated with four different summary length constraints $l$: $15\%$, $20\%$, and $25\%$ as well as $P$, where $P$ is the averaged proportion of the important frames labeled by human annotators. 
The results are shown in Fig.~\ref{fig:quality_eg} (b). When the video summarization network is allowed to select as many key frames as the human annotators, \emph{i.e.} $l=P$, the F1 scores increase significantly. Compared to our unsupervised model $\text{3DST-UNet}_{\text{unsup}}$, the supervised model $\text{3DST-UNet}_{\text{sup}}$ benefits even more from the relaxed summary length constrains.

\begin{figure*}[t]
\begin{centering}
\center\includegraphics[width=1.8\columnwidth]{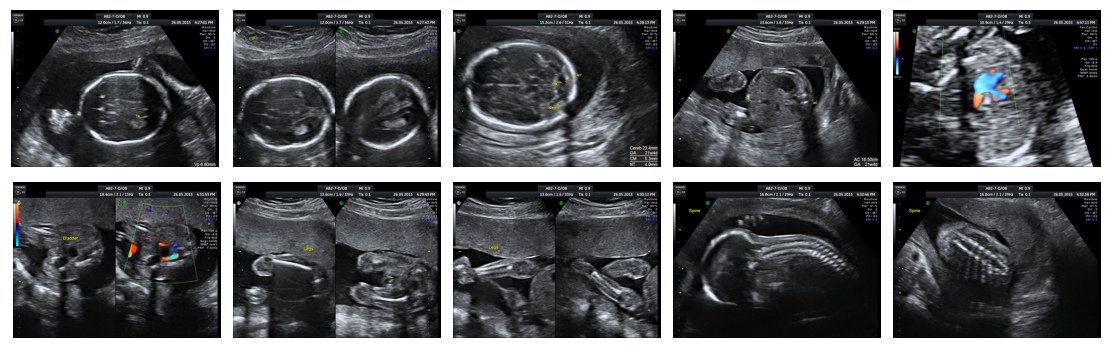}
\par
\end{centering}
\caption{Example key frames for an ultrasound examination recording videos, including the ultrasound plane of brain, abdominal, kidneys, femur, cardiac view and profile and spine. The key frames are acquired during routine ultrasound screenings according to the guidelines in the UK NHS FASP handbook.
}
\label{fig:us_quality_eg}
\end{figure*}

\begin{figure}[t]
\begin{centering}
\center\includegraphics[width=0.83\columnwidth]{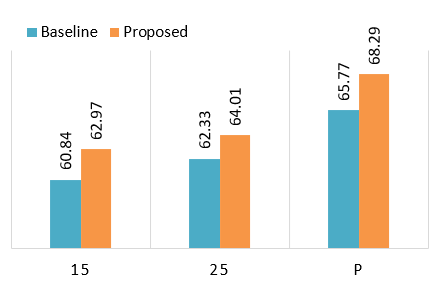}
\par
\end{centering}
\caption{Comparison of performance of the baseline method \cite{RLvSumm_AAAI2018} and the proposed method on medical video summarization. The bar plot gives F1 scores with three different summarization length constrains: $l=15\%, 25\%$ and $P$, where $P$ equals to proportion of the ground-truth important frames for each ultrasound video.
}
\label{fig:ifind_video}
\end{figure}

\subsection{Application to Medical Video Summarization}
For general videos, different individuals can have very different and subjective views regarding to the importance of video segments. The subjectiveness of human-annotators may lead to difficulties in the evaluation of general video summarization tasks. The common solution is to take the average \cite{TVSUM} or maximum F1-scores \cite{SumMe_ECCV2014} over the number of human created summaries. This would not become a problem for medical video summarization where the importance of video frames explicitly follow the clinical diagnostic standards.

We take a fetal ultrasound screening videos dataset to evaluate our method on the medical video summarization task. 
We show that our method is superior to alternative video summarization methods and that it preserves essential information required by clinical diagnostic standards. 
Some example summary frames for ultrasound examination recordings are given in Fig. \ref{fig:us_quality_eg}.

We have compared our ultrasound video summarization results with the baseline method ~\cite{RLvSumm_AAAI2018} with different summary lengths. Fig.~\ref{fig:ifind_video} gives the F1 scores for three different summarization length constrains: again $l=15\%$, $l=25\%$, and $l=P$ where $P$ is the proportion of important frames in the ground-truth for each video in the ultrasound dataset.
As we can see, we outperform \cite{RLvSumm_AAAI2018} for ultrasound video summarization. Our results are even on-par with that of~\cite{liu_MICCAI_2020} (\emph{i.e.}, $63.29$ with $l=15\%$ as given in the paper) whose performance is achieved by using an additional supervised reward term that is informed by manually labeled diagnostic view planes. 

In practical, the proposed video summarization method has the potential to save storage costs of ultrasound screening videos as well as to increase efficiency when browsing patient video data during retrospective analysis or audit without losing essential information.

\subsection{Qualitative Results}

We compare the qualitative video summarization results of the proposed 3DST-UNet method and our baseline method DSN~\cite{RLvSumm_AAAI2018}. In Fig.~\ref{fig:quality_eg}, the ground-truth frame-level importance scores of the video are indicated as light-green bars. The orange bars mark the intervals that have been selected by the summarization methods. Alongside, we show the selected key frames sampled which are the ones that have the highest prediction scores within a video shot. 

We observe that our method preserves the temporal story of the videos by extracting intervals from different sections while focusing on key scenes. This implies that our method is able to preserve information essential for generating meaningful summaries. 


\section{Conclusions}
\label{sec:conclusions}

In this paper we have explored spatio-temporal CNN features for video frame representation in combination with 3DST-UNet, a spatio-temporal 3D U-Net, to model both the spatial and the temporal dependencies amongst video frames. We show with our experiments that 3D CNN features have stronger representation abilities than commonly used spatial 2D CNN features.
The downstream RL framework learns from spatio-temporal latent spaces to predict actions for keeping/rejecting a video frame as being important for a summary. An RL agent takes the latent probability output from a spatio-temporal model to apply learned policies, which can maximize the accumulated reward of actions. We provide a comprehensive ablation study to investigate the contribution of the individual components and critically analyze the impact of hard-coded summary length constraints. The 3DST-UNet-RL model achieves competitive state-of-the-art performance on both general video summarization tasks and a medical video summarization task. The ultrasound video summarization method can be used for a variety of applications also when clinical annotations are unavailable. The proposed framework has the potential to save storage costs as well as to increase efficiency when browsing patient medical video data during retrospective analysis. 

\section*{Acknowledgment}

We thank the volunteers and sonographers from routine fetal screening at St. Thomas' Hospital London. This work was supported by the Wellcome Trust IEH Award [102431] and EPSRC EP/S013687/1. The research was funded/supported by the National Institute for Health Research (NIHR) Biomedical Research Center based at Guy's and St Thomas' NHS Foundation Trust, King's College London and the NIHR Clinical Research Facility (CRF) at Guy's and St Thomas'. Data access only in line with the informed consent of the participants, subject to approval by the project ethics board and under a formal Data Sharing Agreement. The views expressed are those of the author(s) and not necessarily those of the NHS, the NIHR or the Department of Health.


\bibliographystyle{IEEEtran}
\bibliography{vSumm}

\end{document}